\documentclass{article} 
\usepackage{iclr2016_conference,times}
\usepackage{scrextend}
\usepackage{hyperref}
\usepackage{url}
\usepackage{amsmath, amssymb}
\usepackage{graphicx}
\usepackage{pgfplots}
\usepackage{multirow}
\usepackage{epstopdf}
\usepackage{footnote}
\usepackage{color}
\usepackage{algorithm}
\usepackage{algpseudocode}

\makeatletter
\def\BState{\State\hskip-\ALG@thistlm}
\makeatother

\definecolor{cyan}{rgb}{0.0, 0.65, 0.85}
\definecolor{magenta}{rgb}{0.85, 0.15, 0.5}
\definecolor{orange}{rgb}{1.0, 0.5, 0.0}

\graphicspath{{fig/}}
\DeclareGraphicsExtensions{.eps}

\title{Online Sequence Training\\of Recurrent Neural Networks\\with Connectionist Temporal Classification}

\author{Kyuyeon~Hwang \& Wonyong~Sung\\
Department of Electrical and Computer Engineering\\
Seoul National University\\
Seoul, 08826 Korea\\
\texttt{kyuyeon.hwang@gmail.com}\\
\texttt{wysung@snu.ac.kr}\\
}

\begin{document}

\maketitle

\begin{abstract}
Connectionist temporal classification (CTC) based supervised sequence training of recurrent neural networks (RNNs) has shown great success in many machine learning areas including end-to-end speech and handwritten character recognition. For the CTC training, however, it is required to unroll (or unfold) the RNN by the length of an input sequence. This unrolling requires a lot of memory and hinders a small footprint implementation of online learning or adaptation. Furthermore, the length of training sequences is usually not uniform, which makes parallel training with multiple sequences inefficient on shared memory models such as graphics processing units (GPUs). In this work, we introduce an expectation-maximization (EM) based online CTC algorithm that enables unidirectional RNNs to learn sequences that are longer than the amount of unrolling. The RNNs can also be trained to process an infinitely long input sequence without pre-segmentation or external reset. Moreover, the proposed approach allows efficient parallel training on GPUs. For evaluation, phoneme recognition and end-to-end speech recognition examples are presented on the TIMIT and Wall Street Journal (WSJ) corpora, respectively. Our online model achieves 20.7\% phoneme error rate (PER) on the very long input sequence that is generated by concatenating all 192 utterances in the TIMIT core test set. On WSJ, a network can be trained with only 64 times of unrolling while sacrificing 4.5\% relative word error rate (WER).
\end{abstract}

\section{Introduction}

Supervised sequence learning is a regression task where the objective is to learn a mapping function from the input sequence $\mathbf x$ to the corresponding output sequence $\mathbf z$ for all $(\mathbf x, \mathbf z) \in S$ with the given training set $S$, where $\mathbf x$ and $\mathbf z$ can have different lengths. When combined with recurrent neural networks (RNNs), supervised sequence learning has shown great success in many applications including machine translation \citep{bahdanau2014neural, sutskever2014sequence, cho2014learning}, speech recognition \citep{graves2013speech, graves2014towards, hannun2014deepspeech, bahdanau2015end, chorowski2015attention, chan2015listen}, and handwritten character recognition \citep{graves2008unconstrained, frinken2012novel}. Although several attention-based models have been introduced recently, connectionist temporal classification (CTC) \citep{graves2006connectionist} is still one of the most successful techniques in practice, especially for end-to-end speech and character recognition tasks \citep{graves2014towards, hannun2014deepspeech, graves2008unconstrained, frinken2012novel}.

The CTC based sequence training is usually applied to bidirectional RNNs \citep{graves2005framewise}, where both the past and the future information is considered for generating the output at each frame. However, the output of the bidirectional RNNs is available after all of the frames in the input sequence are fed into the RNNs because the future information is backward propagated from the end of the sequence. Therefore, the bidirectional RNNs are not suitable for realtime applications such as incremental speech recognition \citep{fink1998incremental}, that require low-latency output from the RNN. On the other hand, unidirectional RNNs only make use of the past information with some performance sacrifice and are suitable for the low-latency applications. Moreover, the CTC-trained unidirectional RNNs do not need to be unrolled (or unfolded) at the test time. It is shown by \citet{graves2012supervised} that CTC can also be employed for sequence training of unidirectional RNNs on a phoneme recognition task. In this case, the unidirectional RNN also learns the suitable amount of the output delay that is required to accurately process the input sequence.

For the CTC training of both unidirectional and bidirectional RNNs, it is required to unroll the RNNs by the length of the input sequence. By unrolling an RNN $N$ times, every activations of the neurons inside the network are replicated $N$ times, which consumes a huge amount of memory especially when the sequence is very long. This hinders a small footprint implementation of online learning or adaptation. Also, this ``full unrolling'' makes a parallel training with multiple sequences inefficient on shared memory models such as graphics processing units (GPUs), since the length of training sequences is usually not uniform, and thus a load imbalance problem occurs. For unidirectional RNNs, this problem can be addressed by concatenating sequences to make a very long stream of sequences, and training the RNNs with synchronized fixed-length unroll-windows over multiple training streams \citep{chen2014efficient, hwang2015single}. However, it is not straightforward to apply this approach to the CTC training, since the standard CTC algorithm requires full unrolling for the backward variable propagation, which starts from the end of the sequence.

In this paper, we propose an expectation-maximization (EM) based online CTC algorithm for sequence training of unidirectional RNNs. The algorithm allows training sequences to be longer than the amount of the network unroll. Moreover, it can be applied to infinitely long input streams with roughly segmented target sequences (e.g. only with the utterance boundaries and the corresponding transcriptions for training an end-to-end speech recognition RNN). Then, the resulting RNN can run continuously without pre-segmentation or external reset. Due to the fixed unroll amount, the proposed algorithm is suitable for online learning or adaptation systems with constrained hardware resource. Furthermore, the approach can directly be combined with the GPU based parallel RNN training algorithm described in \citet{hwang2015single}. For evaluation, we present examples of end-to-end speech recognition on the Wall Street Journal (WSJ) corpus \citep{paul1992design} with continuously running RNNs.\footnote{Further experiments are performed on TIMIT \citep{garofolo1993darpa} in Appendix~\ref{sec:timit}.} Experimental results show that the proposed online CTC algorithm performs comparable to the almost fully unrolled CTC training even with the small unroll amount that is shorter than the average length of the sequences in the training set. Also, the reduced amount of unroll allows more parallel sequences to be trained concurrently with the same memory use, which results in greatly improved training speed on a GPU.

The paper is organized as follows. In Section~\ref{sec:ctc}, the standard CTC algorithm is explained. Section~\ref{sec:online_ctc} contains the definition of the online sequence training problem and proposes the online CTC algorithm. In Section~\ref{sec:continuous}, the algorithm is extended for the continuously running RNNs, which is followed by its parallelization in Section~\ref{sec:parallel}. In Section~\ref{sec:experiments}, the proposed algorithm is evaluated with speech recognition examples. Concluding remarks follow in Section~\ref{sec:conclusion}.

\section{Connectionist temporal classification (CTC)}
\label{sec:ctc}

\begin{figure}[h]
\begin{center}
\includegraphics[height=1.35in]{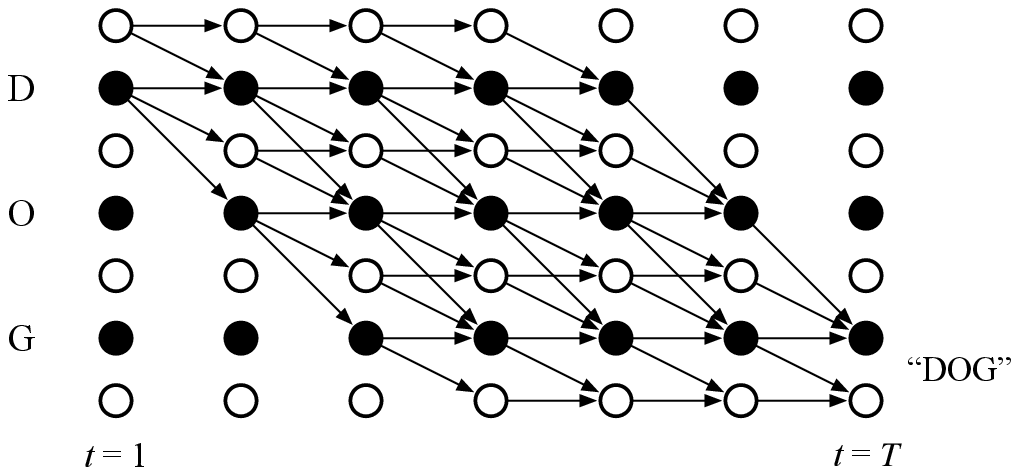}%
\end{center}
\caption{CTC forward-backward example for the target sequence ``DOG'', where the black and white dots represent the labels and blanks, respectively. The arrows indicate the allowed transitions.}
\label{fig:ctc-dog}
\end{figure}

The CTC algorithm \citep{graves2006connectionist, graves2012supervised} considers the order of the output labels of RNNs with ignoring the alignments or timings by introducing an additional blank label, $b$. For the set of target labels, $L$, and its extended set with the additional CTC blank label, $L' = L \cup \{b\}$, the path, $\pi$, is defined as a sequence over $L'$, that is, $\pi \in L'^{T}$, where $T$ is the length of the input sequence, $\mathbf x$. Then, the output sequence, $\mathbf z \in L^{\leq T}$, is represented by $\mathbf z = \mathcal F(\pi)$ with the sequence to sequence mapping function $\mathcal F$. $\mathcal F$ maps any path $\pi$ with the length $T$ into the shorter sequence of the label, $\mathbf z$, by first merging the consecutive same labels into one and then removing the blank labels. Therefore, any sequence of the raw RNN outputs with the length $T$ can be decoded into the shorter labeling sequence, $\mathbf z$, with ignoring timings. This enables the RNNs to learn the sequence mapping, $\mathbf z = \mathcal G(\mathbf x)$, where $\mathbf x$ is the input sequence and $\mathbf z$ is the corresponding target labeling for all $(\mathbf x, \mathbf z)$ in the training set, $S$. More specifically, the gradient of the loss function $\mathcal L(\mathbf x, \mathbf z) = -\ln p(\mathbf z | \mathbf x)$ is computed and fed to the RNN through the softmax layer \citep{bridle1990probabilistic}, of which the size is $|L'|$.

As depicted in \figurename~\ref{fig:ctc-dog}, the CTC algorithm employs the forward-backward algorithm for computing the gradient of the loss function, $\mathcal L(\mathbf x, \mathbf z)$. Let $\mathbf z'$ be the sequence over $L'$ with the length of $2 |\mathbf z| + 1$, where $z'_u=b$ for odd $u$ and $z'_u=z_{u/2}$ for even $u$. Then, the forward variable, $\alpha$, and the backward variable, $\beta$, are initialized by
\begin{align}
\alpha(1, u) =
\begin{cases}
y_b^1 & \text{if } u = 1 \\
y_{z_1}^1 & \text{if } u = 2 \\
0 & \text{otherwise}
\end{cases} \; , \quad
\beta(T, u) =
\begin{cases}
1 & \text{if } u = |\mathbf z'|, |\mathbf z'| - 1 \\
0 & \text{otherwise}
\end{cases},
\label{eq:CTC_init}
\end{align}
where $y_k^t$ is the softmax output of the label $k \in L'$ at time $t$.
The variables are forward and backward propagated as
\begin{align}
\alpha(t, u) = y_{z_u'}^ t \sum_{i=f(u)}^u \alpha(t - 1, i) \; , \quad
\beta(t, u) = \sum_{i=u}^{g(u)} \beta(t + 1, i) y_{z'_i}^{t+i},
\label{eq:recursion}
\end{align}
where
\begin{align}
f(u) =
\begin{cases}
u - 1&\text{if } z'_u=b\text{ or } z'_{u-2}=z'_u\\
u - 2&\text{otherwise}
\end{cases} \; , \quad
g(u) =
\begin{cases}
u + 1&\text{if } z'_u=b\text{ or } z'_{u+2}=z'_u\\
u + 2&\text{otherwise}
\end{cases}
\end{align}
with the boundary conditions:
\begin{align}
\alpha(t, 0) = 0, \; \forall t \; , \quad
\beta(t, |\mathbf z'|+1) = 0, \; \forall t .
\end{align}
Then, the error gradient with respect to the input of the softmax layer at time $t$, $a_k^t$, is computed as
\begin{align}
\frac {\partial \mathcal L(\mathbf x, \mathbf z)} {\partial a_k^t} = y_k^t - \frac {1}{p(\mathbf z | \mathbf x)} \sum_{u \in B(\mathbf z, k)} {\alpha(t, u) \beta(t, u)},
\label{eq:CTC_gradient}
\end{align}
where
$ B(\mathbf z, k) = \{ u : \mathbf z'_u = k\} $ and
$ p(\mathbf z | \mathbf x) = \alpha(T, |\mathbf z'|) + \alpha(T, |\mathbf z'| - 1)
$.


\section{Online sequence training}
\label{sec:online_ctc}

\subsection{Problem definition}

Throughout the paper, the online sequence training problem is defined as follows.
\begin{itemize}
\item The training set $S$ consists of pairs of the input sequence $\mathbf x$ and the corresponding target sequence $\mathbf z$, that is, $(\mathbf x, \mathbf z) \in S$.
\item The estimation model $\mathcal M$ learns the general mapping $\mathbf z = \mathcal G (\mathbf x)$, where the training sequences $(\mathbf x, \mathbf z) \in S$ are sequentially given.
\item For each $(\mathbf x, \mathbf z) \in S$ and at time $t$, only the fraction of the input sequence up to time $t$, $\mathbf x_{1:t}$, and the entire target sequence, $\mathbf z$, are given, where $1 \leq t \leq |\mathbf x|$. The length of the input sequence, $|\mathbf x|$, is unknown except when $t=|\mathbf x|$.
\item The parameters of the estimation model $\mathcal M$ is updated in the manner of online learning, that is, they can be frequently updated even before seeing the entire input sequence $\mathbf x$.
\end{itemize}

This online learning problem usually occurs in real world when a human learns a language from texts and the corresponding audio. For example, when watching movies with subtitles, we are given the entire target sequence (subtitle for the current utterance) and the input sequence (the corresponding audio) up to the current time, $t$. We cannot access the future audio and even do not know exactly when the utterance will end (at $t=|\mathbf x|$).

When RNNs are trained with the standard CTC algorithm, it is difficult to determine how much amount of unrolling is needed before the entire sequence $\mathbf x$ is given, since the length of $\mathbf x$ is unknown at time $t < |\mathbf x|$. Also, it is not easy to learn the sequences that are longer than the unroll amount, which is often constrained by the hardware resources.

\subsection{Overview of the proposed approach}

\begin{figure}[h]
\begin{center}
\includegraphics[width=5.0in]{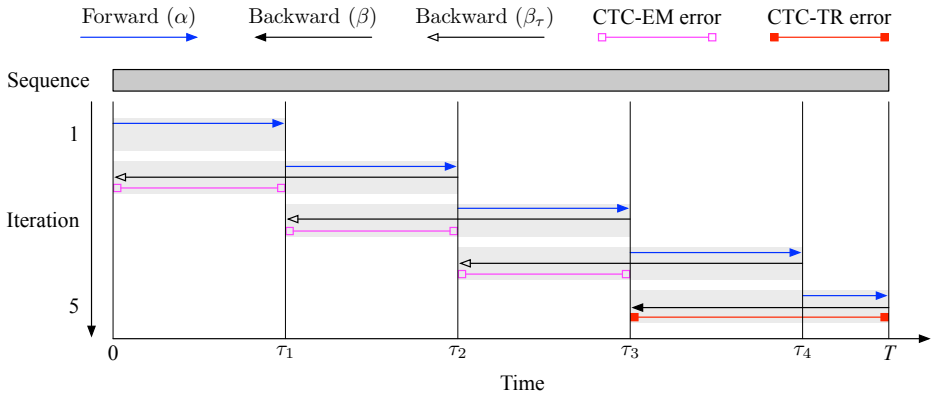}%
\end{center}
\caption{Online CTC($2h'$;~$h'$) algorithm depicted for a single sequence that is longer than the RNN unroll amount. The shaded areas indicate the range of the RNN unrolling at each iteration.}
\label{fig:online}
\end{figure}

We propose an online CTC algorithm where the RNN can learn the sequences longer than the unroll amount, $h$. The algorithm is based on the truncated backpropagation through time (BPTT) algorithm \citep{werbos1990backpropagation} with the forward step size of $h'$ and the unroll amount of $h$, which is called BPTT($h$;~$h'$), as proposed in \citet{williams1990efficient}. Algorithm~\ref{algo:BPTT} describes the BPTT($h$;~$h'$) algorithm combined with the CTC loss, where $T$ is the length of the training sequence, $\mathbf x$

\begin{algorithm}
\caption{Online CTC training with BPTT($h$;~$h'$) for a single sequence}\label{algo:BPTT}
\begin{algorithmic}[1]
\Procedure{BPTT($h$;~$h'$)}{}
\State $\tau_0 \gets 0$
\State $n \gets 1$
\While {$\tau_{n-1} < T$}
\State $\tau'_n \gets \max\{1, nh' - h + 1\}$ \Comment{Start index of unrolling}
\State $\tau_n \gets \min\{nh', T\}$ \Comment{End index of unrolling}
\State RNN forward activation from $t=\tau_{n-1} + 1$ to $\tau_n$
\State CTC($h$;~$h'$) error computation on the softmax output layer \Comment{Algorithm~\ref{algo:onlineCTC}}
\State RNN backward error propagation from $t=\tau_{n}$ to $\tau'_{n}$
\State RNN gradient computation and weight update
\State $n \gets n + 1$
\EndWhile
\EndProcedure
\end{algorithmic}
\end{algorithm}

However, although BPTT($h$;~$h'$) is designed for online training of RNNs, employing the standard CTC loss function requires full unrolling of the networks. Therefore, we propose the CTC($h$;~$h'$) algorithm for computing the CTC loss in the online manner as in BPTT($h$;~$h'$) as in Algorithm~\ref{algo:onlineCTC}. The algorithm is also depicted in \figurename~\ref{fig:online} with the example in which the length of the sequence, $T = |\mathbf x|$, is 2.5 times as long as the unroll amount.

\begin{algorithm}
\caption{CTC($h$;~$h'$) error computation at the iteration $n$}\label{algo:onlineCTC}
\begin{algorithmic}[1]
\Procedure{CTC($h$;~$h'$)}{}
\State $\tau_{n-1} \gets (n-1)h'$
\State $\tau'_n \gets \max\{1, nh' - h + 1\}$
\State $\tau'_{n+1} \gets \max\{1, (n+1)h' - h + 1\}$
\State $\tau_n \gets \min\{nh', T\}$
\If {$n = 1$}
\State Initialize the CTC forward variable, $\alpha$, at $t=1$ \Comment{\eqref{eq:CTC_init}}
\EndIf
\State CTC forward propagation of $\alpha$ from $t=\tau_{n-1} + 1$ to $\tau_n$ \Comment{\eqref{eq:recursion}}
\If {$\tau_n = T$} \Comment{CTC-TR in Section~\ref{ssec:ctc-tr}}
\State Initialize the CTC-TR backward variable, $\beta$, at $t=T$ \Comment{\eqref{eq:CTC_init}}
\State CTC-TR backward propagation of $\beta$ from $t=T$ to $\tau'_n$ \Comment{\eqref{eq:recursion}}
\State CTC-TR error computation with $\alpha$ and $\beta$ on $t \in [\tau'_n, T]$ \Comment{\eqref{eq:CTC_gradient}}
\Else \Comment{CTC-EM in Section~\ref{ssec:ctc-em}}
\State Initialize the CTC-EM backward variable, $\beta_{\tau_{n}}$, at $t=\tau_{n}$ \Comment{\eqref{eq:CTC-EM_init}}
\State CTC-EM backward propagation of $\beta_{\tau_{n}}$ from $t=\tau_{n}$ to $\tau'_n$ \Comment{\eqref{eq:recursion}}
\State CTC-EM error computation with $\alpha$ and $\beta_{\tau_{n}}$ on $t \in [\tau'_n, \tau'_{n+1}-1]$ \Comment{\eqref{eq:CTC_gradient}}
\State Set error to zero on $t \in [\tau'_{n+1}, \tau_{n}]$
\EndIf
\EndProcedure
\end{algorithmic}
\end{algorithm}


CTC($h$;~$h'$) consists of two CTC algorithms. The first one is the truncated CTC (CTC-TR), which is basically the standard CTC algorithm applied at the last iteration with truncation. In the other iterations, the generalized EM based CTC algorithm (CTC-EM) is employed from $t=\max\{1, nh' - h + 1\}$ to $\max\{0, (n + 1)h' - h\}$ with the modified backward variable, $\beta_\tau$. The CTC-TR and CTC-EM algorithms are explained in Section~\ref{ssec:ctc-tr} and Section~\ref{ssec:ctc-em}, respectively. Note that simply setting $h=2h'$ works well in practice. In this setting, we denote the algorithm as CTC($2h'$;~$h'$).

\subsection{CTC-TR: Standard CTC with truncation}
\label{ssec:ctc-tr}

With the standard CTC algorithm, it is not possible to compute the backward variables when $\tau_n < T$, as the future information beyond $\tau_n$ cannot be accessed. Therefore, we only compute the CTC errors at the last iteration, where $\tau_n = T$ as in Algorithm~\ref{algo:onlineCTC}. In this case, however, the gradients are only available in the unroll range. Since the backward propagation is truncated at the beginning of the unroll range, we call the CTC algorithm in this range as truncated CTC, or CTC-TR. Also, we call the range that is covered by the CTC-TR algorithm as the CTC-TR coverage.

The RNN can be trained only with CTC-TR if there are sufficient labels that occur within the CTC-TR coverage. However, the CTC-TR coverage decreases by making the unroll amount smaller. Then, the percentage of the effective training frames, which actually generate the output errors, goes down, and the efficiency of training decreases. Also, the effective size of the training set gets smaller, which results in the loss of the generalization performance of the RNN. Therefore, for maintaining the training performance while reducing the unroll amount, it is critical to make use of the full training frames by employing the CTC-EM algorithm, which is described in Section~\ref{ssec:ctc-em}.

\subsection{CTC-EM: Online CTC with expectation-maximization (EM)}
\label{ssec:ctc-em}

\begin{figure}[h]
\begin{center}
\includegraphics[height=1.35in]{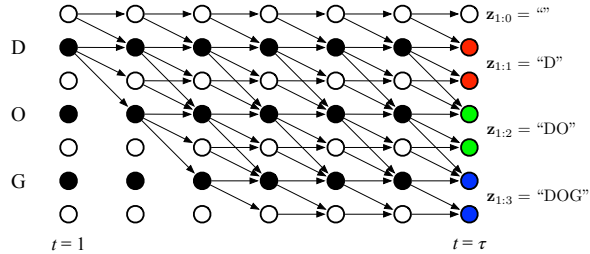}%
\end{center}
\caption{Forward-backward algorithm of CTC-EM for the target sequence ``DOG'', where the black and white dots represent the labels and CTC blanks, respectively. The arrows represent the paths with allowed transitions.}
\label{fig:online-dog}
\end{figure}

Assume that only the fraction of the input sequence, $\mathbf x_{1:\tau}$, is available. Then, as shown in the \figurename~\ref{fig:online-dog}, there are $|\mathbf z| + 1$ possible partial labelings.\footnote{Although $\mathbf z_{1:m}$ is not possible by the standard CTC formulation when $m > \tau$, we can still say that $\mathbf z_{1:m}$ is a possible labeling with a probability of zero without loss of generality.} Let $\mathbf z_{1:m}$ be the subsequence of $\mathbf z$ with the first $m$ labels. Also we define $Z$ as the set that consists of these labeling sequences:
\begin{align}
Z = \{\mathbf z_{1:m} : 0 \leq m \leq |\mathbf z|\}.
\end{align}
One of the most simple approach for training the network under this condition is to choose the most likely partial alignment from $Z$ and compute the standard CTC error by regarding the partial alignment as the ground truth labeling. For example, we can select $\mathbf z_{1:m'}$ where $m'=\operatornamewithlimits{arg\,max}_m \alpha(\tau, m)$ since $\alpha(\tau, m)$ is a posterior probability $p(\mathbf z_{1:m} | \mathbf x_{1:\tau}, \mathbf w^{(n)})$ with the current network parameter $\mathbf w^{(n)}$. This is a well-known hard-EM approach. This simple idea can easily be extended to the more sophisticated soft-EM approach as follows. First, select one of the partial labelings in $Z$ with the probability $p(\mathbf z_{1:m} | Z, \mathbf x_{1:\tau}, \mathbf w^{(n)})$ estimated by the RNN with current parameters (E-step). Then, maximize the probability of that labeling by adjusting the parameters (M-step).

This optimization problem is readily reduced into the generalized EM algorithm. Specifically, the expectation step is represented as
\begin{align}
Q_\tau(\mathbf w | \mathbf x, \mathbf z, \mathbf w^{(n)}) &=
\mathbb{E}_{\mathbf z_{1:m} | Z, \mathbf x_{1:\tau}, \mathbf w^{(n)}}
\left[ \ln p(\mathbf z_{1:m} | \mathbf x_{1:\tau}, \mathbf w)  \right] \\
&=
\sum_{m=0}^{|\mathbf z|}
p(\mathbf z_{1:m} | Z, \mathbf x_{1:\tau}, \mathbf w^{(n)})
\ln p(\mathbf z_{1:m} | \mathbf x_{1:\tau}, \mathbf w),
\end{align}
where $\mathbf w^{(n)}$ is the set of the network parameters at the current iteration, $n$. In the maximization step of the generalized EM approach, we try to maximize $Q_\tau$ by finding new parameters $\mathbf w^{(n+1)}$ that satisfies $Q_\tau(\mathbf w^{(n+1)} | \mathbf x, \mathbf z, \mathbf w^{(n)}) \geq Q_\tau(\mathbf w^{(n)} | \mathbf x, \mathbf z, \mathbf w^{(n)})$. As proved in Appendix~\ref{sec:ctc-em_derivation}, this is equivalent to the optimization problem where the objective is to minimize the loss function defined as $\mathcal{L_\tau (\mathbf x, \mathbf z)} = -\ln p(Z | \mathbf x_{1:\tau})$. Then, the gradient of the loss function with respect to the input of the softmax layer is
\begin{align}
\frac{ \partial \mathcal{L_\tau (\mathbf x, \mathbf z)}}
{\partial a_k^t}
=
y_k^t - \frac {1}{p(Z | \mathbf x_{1:\tau})} \sum_{u \in B(\mathbf z, k)} {\alpha(t, u) \beta_\tau(t, u)},
\end{align}
where $p(Z | \mathbf x_{1:\tau})$ can be computed by
\begin{align}
p(Z | \mathbf x_{1:\tau})=
\sum_{u=1}^{|\mathbf z'|}
\alpha(\tau, u)
\end{align}
and the backward variable, $\beta_\tau (t, u)$, is initialized as
\begin{align}
\beta_{\tau}(\tau, u) = 1, \; \forall u .
\label{eq:CTC-EM_init}
\end{align}
The new backward variable is backward propagated using the same recursion in \eqref{eq:recursion}, and the error gradients are computed with \eqref{eq:CTC_gradient} as in the standard CTC algorithm. See Appendix~\ref{sec:ctc-em_derivation} for the derivation of the above equations.


\section{Training continuously running RNNs}
\label{sec:continuous}

\begin{figure}[h]
\begin{center}
\includegraphics[width=5.0in]{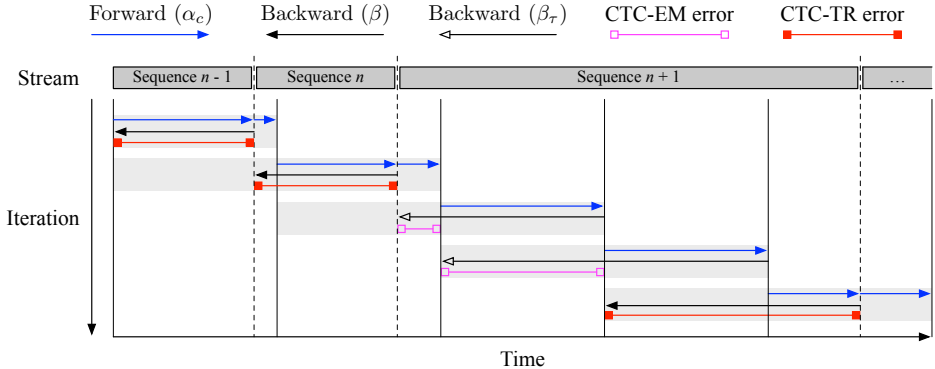}%
\end{center}
\caption{Online CTC training with a continuous stream of sequences. The shaded areas indicate the range of RNN unrolling at each iteration.}
\label{fig:online-cont}
\end{figure}

In this section, the proposed online CTC algorithm in Section~\ref{sec:online_ctc} is extended for training infinitely long streams. The training stream can be naturally very long with the target sequence boundaries, or can be generated by concatenating training sequences in a certain order. When trained on this training stream without external reset of the RNN at the sequence boundaries, the resulting RNN can also continuously process infinitely long input streams without pre-segmentation or external reset. This property is useful for realtime speech recognition or keyword spotting (spoken term detection) systems since we can remove the frontend voice activity detector \citep{sohn1999statistical} for detecting and pre-segmenting utterances.

The CTC($h$;~$h'$) algorithm can directly be applied to the infinitely long training streams as shown in \figurename~\ref{fig:online-cont}. When the sequence boundaries are reached during the forward activation, we perform CTC-TR, initialize the forward variable, and process the next sequence with some frame offset. Also, care should be taken on the transition of CTC labels at the boundary. Assume that the last label of the sequence $n$ and the first label of the sequence $n+1$ are the same. Then, a CTC blank label should be inserted between two sequences since the same labels that occur consecutively in the decoding path are folded into one label. In practice, this folding can easily be prevented by forcing the blank label at the first frame of each sequence by modifying the initialization of the forward variable as follows:
\begin{align}
\alpha_c(1, u) =
\begin{cases}
y_b^1 & \text{if } u = 1 \\
0 & \text{otherwise}
\end{cases},
\end{align}
where the subscript $c$ indicates the continuous CTC training.

\section{Parallel training}
\label{sec:parallel}

In a massively parallel shared memory model such as a GPU, efficient parallel training is achieved by making use of the memory hierarchy. For example, computing multiple frames together reduces the number of read operations of the network parameters from the slow off-chip memory by temporarily storing them on the on-chip cache memory and reuse them multiple times. For training RNNs on a GPU, this parallelism can be explored by employing multiple training sequences concurrently \citep{hwang2015single}.

The continuous CTC($h$;~$h'$) algorithm in Section~\ref{sec:continuous} can be directly extended for parallel training with multiple streams. Since the forward step size and the unroll amount is fixed, the RNN forward, backward, gradient computation, and weight update steps can be synchronized over multiple training streams. Thus, the GPU based parallelization approach in \citet{hwang2015single} can be employed for the RNN training. Although the computations in the CTC($h$;~$h'$) algorithm are relatively fewer than those of the RNN, further speed-up can be achieved by parallelizing the CTC algorithm similarly.

\section{Experiments}
\label{sec:experiments}

\subsection{End-to-end speech recognition with RNNs}

For the evaluation of the proposed approach, we present examples of character-level speech recognition with end-to-end trained RNNs without external language models. The speech recognition RNN is similar to the one in \citet{graves2014towards} except that our model employs unidirectional long short-term memory (LSTM) network \citep{hochreiter1997long} trained with the online CTC algorithm on the continuous stream of speech, instead of the bidirectional LSTM network with the sentence-wise CTC training.

Specifically, the experiments are performed with the deep unidirectional LSTM network with 3 LSTM layers, where each layer has 768 LSTM cells. The output layer is a 31-dimensional softmax layer. Each unit of the softmax layer represents one of the posterior probabilities of 26 alphabet characters, two special characters (. and '), a whitespace character, the end of sentence (EOS) symbol, and the CTC blank label. The input of the network is a 123-dimensional vector that consists of a 40-dimensional log Mel-frequency filterbank feature vector plus energy, and their delta and delta-delta values. The feature vectors are extracted from the speech waveform in every 10 ms with 25 ms Hamming window using HTK \citep{young1997htk}. Before being fed into the RNN, feature vectors are element-wisely normalized to the zero mean and the unit standard deviation, where the statistics are extracted from the training set.

\subsection{Wall Street Journal (WSJ) corpus}

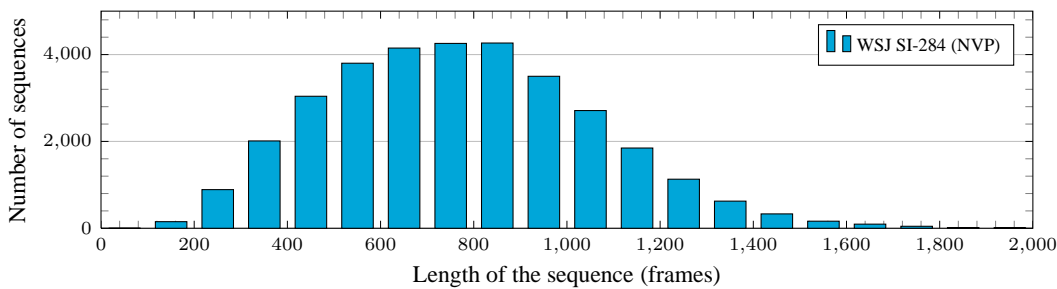
\begin{figure}[h]
\centering
\centerline{%
\begin{tikzpicture}
\begin{axis}
[
width=\columnwidth,
height=0.32\columnwidth,
bar width=0.03\columnwidth,
compat=1.3,
xmin=0,
ymin=0,
xmax=2000,
ymax=5000,
label style={font=\footnotesize},
xlabel={Length of the sequence (frames)},
ylabel={Number of sequences},
xlabel shift=-2pt,
ylabel shift=-2pt,
ybar legend,
legend style={font=\scriptsize,at={(0.98,0.94)},anchor=north east},
tick label style={font=\scriptsize},
ymajorgrids,
minor x tick num=4,
minor y tick num=4,
xtick pos=both,
xtick align=inside,
major tick style={line width=0.010cm, black},
major tick length=0.10cm
]%
\legend{WSJ SI-284 (NVP)};
\addplot[ybar, fill=cyan]
file{data/hist_si284.txt};
\end{axis}%
\end{tikzpicture}%
}%
\caption{Histogram of the length of the sequences in the WSJ SI-284 training set, where only the utterances with non-verbalized punctuations (NVPs) are considered. The feature frames are extracted with the period of 10 ms.}
\label{fig:hist_wsj}
\end{figure}

The experiments are performed on the Wall Street Journal (WSJ) \citep{paul1992design} corpus. For the training, the subset of the WSJ SI-284 set is used, where only the utterances with non-verbalized punctuations (NVPs) are included, resulting in about 71 hours of utterances. The histogram of the length of the sequences in the training set is shown in \figurename~\ref{fig:hist_wsj}. Note that the average length of the sequences is 772.5 frames. If we unroll the network over $N$ frames, the sequences longer than $N$ frames will not be fully covered by CTC-TR.

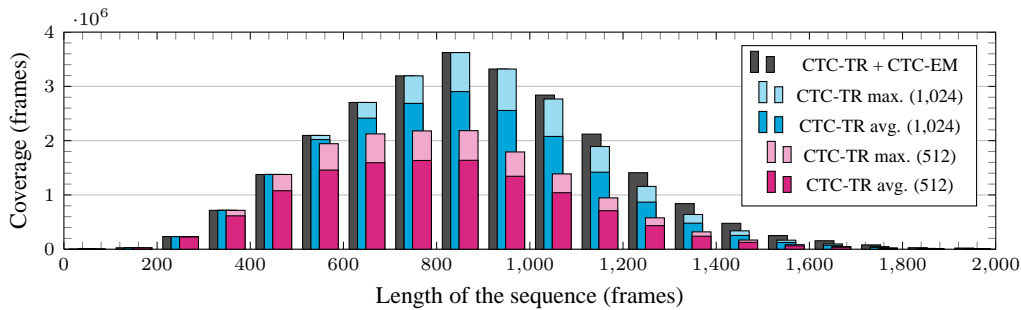
\begin{figure}[h]
\centering
\centerline{%
\begin{tikzpicture}
\begin{axis}
[
width=\columnwidth,
height=0.32\columnwidth,
compat=1.3,
bar width=0.018\columnwidth,
xmin=0,
ymin=0,
xmax=2000,
ymax=4e6,
label style={font=\footnotesize},
xlabel={Length of the sequence (frames)},
ylabel={Coverage (frames)},
xlabel shift=-2pt,
ylabel shift=-2pt,
ybar legend,
legend style={font=\scriptsize,at={(0.98,0.94)},anchor=north east},
tick label style={font=\scriptsize},
domain=1:512,
ymajorgrids,
minor x tick num=4,
minor y tick num=4,
xtick pos=both,
xtick align=inside,
major tick style={line width=0.010cm, black},
major tick length=0.10cm
]%
\legend{CTC-TR + CTC-EM, CTC-TR max. ({1,024}), CTC-TR avg. ({1,024}), CTC-TR max. (512), CTC-TR avg. (512)};
\addplot[ybar, fill=black!70, xshift=-0.008\columnwidth]
file{data/hist_si284_frames.txt};
\addplot[ybar, fill=cyan!40, xshift=0]
file{data/hist_si284_frames_tr1024.txt};
\addplot[ybar, fill=cyan, xshift=0]
file{data/hist_si284_frames_tr1024_avg.txt};
\addplot[ybar, fill=magenta!40, xshift=0.008\columnwidth]
file{data/hist_si284_frames_tr512.txt};
\addplot[ybar, fill=magenta, xshift=0.008\columnwidth]
file{data/hist_si284_frames_tr512_avg.txt};
\end{axis}%
\end{tikzpicture}%
}%
\caption{Coverage of the trainable frames with respect to the length of the sequences in the WSJ SI-284 (NVP) training set. The average and maximum coverages of CTC-TR on continuous training streams are visualized for the unroll amount of 512 and 1,024 when CTC($2h'$;~$h'$) is applied. Note that the proposed online CTC algorithm (CTC-TR + CTC-EM) covers the entire training frames.}
\label{fig:coverage_wsj}
\end{figure}

In \figurename~\ref{fig:coverage_wsj}, the CTC-TR coverage is further analyzed with respect to the length of the sequence and the unroll amount. When the stream of sequences are trained with the continuos CTC algorithm, the CTC-TR coverage varies depending on the frame offsets of CTC($h$;~$h'$). The average coverage is calculated assuming that the offset is uniformly distributed. If the probability that a certain frame is included in the coverage is greater than zero, then the frame is included in the maximum coverage. For the experiments, we only consider CTC($2h'$;~$h'$), that is, the unroll amount is twice as much as the forward step size. Then, unrolling the network 1,024 times results in the CTC-TR coverage of 79.48 \% on average and 95.69 \% at maximum. On the other hand, when the unroll amount is 512, CTC-TR only covers 48.16 \% on average and 63.27 \% at maximum. Note that the full coverage is achieved when CTC-TR is combined with CTC-EM.

The WSJ Nov'93 20K development set and the WSJ Nov'92 20K evaluation set are used as the development (validation) set and the test (evaluation) set, respectively. For the final evaluation of the network after training, a single test stream is used that is generated by concatenating all of the 333~utterances in the test set.

\subsection{Training procedure}

The networks are trained on a GPU as in Section~\ref{sec:parallel} with the memory usage constraint. To maintain the memory usage same while changing the unroll amount, we fixed the total amount of unrolling over multiple training streams to 16,384. For example, the number of parallel streams become 8 with the unroll amount of 2,048 and 32 with 512 times of unrolling. The total amount of GPU memory usage is about 9.5 GiB in our implementation based on \citet{hwang2015single}.

The performance evaluation of the network is performed at every 10,485,760 training frames (i.e. $N$ continuous training streams with the length of $10,485,760/N$ each) in terms of word error rate (WER) on the 128 parallel development streams of which length is 16,384 each. For this intermediate evaluation, best path decoding \citep{graves2006connectionist} is employed for fast computation.

For the online update of the RNN parameters, the stochastic gradient descent (SGD) method is employed and accelerated by the Nesterov momentum of 0.9 \citep{nesterov1983method, bengio2013advances}. Also, the network is annealed by combining the early stopping technique as follows. If the network performance based on the intermediate evaluation is not improved for 11 consecutive times (10 times of retry), the learning rate is reduced by the factor of 10 and the training is resumed from the second best network. The training starts from the learning rate of $10^{-5}$ and finishes when the learning rate becomes less than $10^{-7}$.

The pre-trained network is used for CTC-TR and CTC-EM combined training because the expectation step of CTC-EM requires the RNN to align the target labels in a certain level. The pre-trained networks are obtained by early stopping the CTC-TR training of the networks when the performance is not improved during 6 consecutive intermediate evaluations using the learning rate of $10^{-5}$. For the CTC-TR and CTC-EM combined training with the unroll amount of 512, 1,024, and 2,048, the training starts from the pre-trained network that is trained with the same amount of unrolling. Otherwise, for the combined training with the unrolling less than 512 times, we use the pre-trained network with the unroll amount of 512.

\subsection{Evaluation}

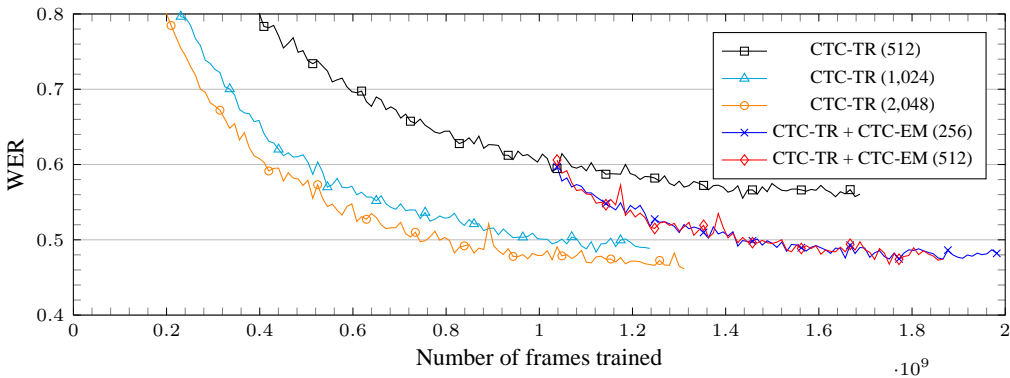
\begin{figure}[h]
\centering
\centerline{%
\begin{tikzpicture}
\begin{axis}
[
width=\columnwidth,
height=0.4\columnwidth,
compat=1.3,
xmin=0.0e9,
ymin=0.4,
xmax=2.0e9,
ymax=0.8,
label style={font=\footnotesize},
xlabel={Number of frames trained},
ylabel={WER},
xlabel shift=-2pt,
ylabel shift=-2pt,
legend style={font=\scriptsize,at={(0.98,0.94)},anchor=north east},
tick label style={font=\scriptsize},
domain=1:512,
ymajorgrids,
minor x tick num=4,
minor y tick num=4,
log basis x={10},
xtick pos=both,
xtick align=inside,
major tick style={line width=0.010cm, black},
major tick length=0.10cm
]%
\legend{CTC-TR (512)\hphantom{,0}, CTC-TR ({1,024}), CTC-TR ({2,048}), CTC-TR + CTC-EM (256), CTC-TR + CTC-EM (512)};
\addplot[color=black, solid, mark=square, mark size=1.5, mark repeat=10,mark options=solid]
file{data/512_tr_1e-05.txt};
\addplot[color=cyan, solid, mark=triangle, mark size=2, mark repeat=10,mark options=solid]
file{data/1024_tr_1e-05.txt};
\addplot[color=orange, solid, mark=o, mark size=1.5, mark repeat=10,mark options=solid]
file{data/2048_tr_1e-05.txt};
\addplot[color=blue, solid, mark=x, mark size=2, mark repeat=10,mark options=solid]
file{data/256_online_1e-05.txt};
\addplot[color=red, solid, mark=diamond, mark size=2, mark repeat=10,mark options=solid]
file{data/512_online_1e-05.txt};
\end{axis}%
\end{tikzpicture}%
}%
\caption{Convergence curves in terms of WER on the development set with the various unroll amounts of 256, 512, 1,024, and 2,048, and the fixed learning rate of $10^{-5}$.}
\label{fig:convergence}
\end{figure}

\figurename~\ref{fig:convergence} shows the convergence curves in terms of WER on the development set with various unroll amounts and training algorithms, where the unroll amount is twice the forward step size. The convergence speed of the CTC-TR only training decreases when the unroll amount becomes smaller. This is because the percentage of the effective training frames become smaller due to the reduced CTC-TR coverage. Also, it can be observed that the performance of the CTC-TR only trained network with 512 times of unrolling converges to the worse WER than those of the other networks due to the reduced size of the effective training set. On the other hand, the convergence curves of the CTC-TR and CTC-EM combined training with the unroll amounts of 256 and 512 are similar to that of the CTC-TR only training with 2,048 times of unrolling. Considering that the average sequence length of the training set is 772.5 frames, the results are quite encouraging.

The evidence of the similar convergence curves with the different unroll amounts implies that the training can be accelerated under the memory usage constraint by employing more parallel training streams with less unrolling. To examine how much speed-up can be achieved on a GPU, further experiments are performed as in \tablename~\ref{tbl:wsj_wer}. The training speed is measured on the system equipped with NVIDIA GeForce Titan X GPU and Intel Xeon E5-2620 CPU. For the final character error rate (CER) and WER report on the test set, the output of the RNN is decoded by the CTC beam search \citep{graves2014towards} without language models. As shown in the table, we can achieve a great amount of speedup without sacrificing much WERs. Also, it is possible to train a network with only 64 times of unrolling, which corresponds to 640 ms window, at the cost of 4.5\% relative WER. See Appendix~\ref{sec:timit} for further experiments on TIMIT \citep{garofolo1993darpa}.

\begin{table}[t]
\caption{Comparison of the CTC-TR coverages, the CER and WERs on the test set, and the training speeds on the GPU with the varying amounts of unrolling}
\label{tbl:wsj_wer}
\begin{center}
\begin{tabular}{l|cc|cc|cc}
\multicolumn{1}{c|}{\bf \# Streams}  &\multicolumn{2}{c|}{\bf CTC-TR coverage (\%)} &\multicolumn{2}{c|}{\bf CER / WER (\%)} &\multicolumn{2}{c}{\bf Speed (frames/s)} \\
\multicolumn{1}{c|}{\bf $\times$ \# Unroll}&\multicolumn{1}{c}{Average} &\multicolumn{1}{c|}{Maximum} &\multicolumn{1}{c}{CTC-TR} &\multicolumn{1}{c|}{+ CTC-EM} &\multicolumn{1}{c}{CTC-TR} &\multicolumn{1}{c}{+ CTC-EM}
\\  \hline\hline \rule{0pt}{2.5ex}%
\hphantom{00}8 $\times$ 2,048 &97.84 &\hphantom{0}99.995 &- &10.6 / 38.4 &\hphantom{0}3.81 k &\hphantom{0}3.80 k \\
\hphantom{0}16 $\times$ 1,024 &79.48 &95.69 &11.2 / 39.1 &10.9 / 38.6 &\hphantom{0}6.79 k &\hphantom{0}6.60 k \\
\hphantom{0}32 $\times$ 512 &48.16 &63.27 &13.9 / 47.2 &10.9 / 38.8 &12.58 k &11.70 k \\
\hphantom{0}64 $\times$ 256 &24.82 &33.06 &- &11.2 / 39.7 &18.03 k &15.99 k \\
128 $\times$ 128 &12.43 &16.57 &- &11.3 / 40.0 &23.64 k &20.54 k \\
256 $\times$ 64 &\hphantom{0}6.21 &\hphantom{0}8.29 &- &11.4 / 40.1 &26.98 k &22.24 k 
\end{tabular}
\end{center}
\end{table}

\section{Concluding remarks}
\label{sec:conclusion}

Throughout the paper, the online CTC($h$;~$h'$) algorithm is proposed for online sequence training of unidirectional RNNs. The algorithm consists of CTC-TR and CTC-EM. CTC-TR is the standard CTC algorithm with truncation and CTC-EM is the generalized EM based algorithm that covers the training frames that CTC-TR cannot be applied. The proposed algorithm requires the unroll amount less than the length of the training sequence and is suitable for small footprint online learning systems or massively parallel implementation on a shared memory model such as a GPU. Also, the online CTC algorithm is extended for training continuously running RNNs without external reset, and evaluated in the TIMIT experiments with a continuous input speech. On the WSJ corpus, the experimental results indicate that when the memory capacity is constrained, the proposed approach achieves significant speed-up on a GPU without sacrificing the performance of the resulting RNN much. We expect that further acceleration of training will be possible with lower performance loss when different unroll amounts are used in the pre-training, main training, and annealing stages.

\subsubsection*{Acknowledgments}


This work was supported in part by the Brain Korea 21 Plus Project and the National Research Foundation of Korea (NRF) grant funded by the Korea government (MSIP) (No.~2015R1A2A1A10056051).

\bibliography{refs}
\bibliographystyle{iclr2016_conference}


\vfill
\pagebreak

\appendix
\section*{Appendices}
\addcontentsline{toc}{section}{Appendices}
\renewcommand{\thesubsection}{\Alph{subsection}}

\subsection{Derivation of the CTC-EM equations}
\label{sec:ctc-em_derivation}

In the maximization step, the objective is to obtain the derivative of $Q_\tau(\mathbf w | \mathbf x, \mathbf z, \mathbf w^{(n)})$ with respect to the input of the softmax layer, $a_k^t$, at time $t$. We first differentiate $Q_\tau$ with respect to $y_k^t$ at $\mathbf w=\mathbf w^{(n)}$:
\begin{align}
\left. \frac{ \partial Q_\tau(\mathbf w | \mathbf x, \mathbf z, \mathbf w^{(n)}) } { \partial y_k^t } \right| _{\mathbf w = \mathbf w^{(n)}}  &=
\sum_{m=0}^{|\mathbf z|}
p(\mathbf z_{1:m} | Z, \mathbf x_{1:\tau}, \mathbf w^{(n)})
\frac{ \partial \ln p(\mathbf z_{1:m} | \mathbf x_{1:\tau}, \mathbf w^{(n)}) } { \partial y_k^t }.
\label{eq:m_deriv}
\end{align}
With Bayes' rule, we obtain
\begin{align}
p(\mathbf z_{1:m} | Z, \mathbf x_{1:\tau}, \mathbf w^{(n)}) =
\frac{p(\mathbf z_{1:m}, Z | \mathbf x_{1:\tau}, \mathbf w^{(n)})}
{p(Z | \mathbf x_{1:\tau}, \mathbf w^{(n)})}
=
\frac{p(\mathbf z_{1:m} | \mathbf x_{1:\tau}, \mathbf w^{(n)})}
{p(Z | \mathbf x_{1:\tau}, \mathbf w^{(n)})},
\end{align}
and with simple calculus,
\begin{align}
\frac{ \partial \ln p(\mathbf z_{1:m} | \mathbf x_{1:\tau}, \mathbf w^{(n)}) } { \partial y_k^t } =
\frac{1}{ p(\mathbf z_{1:m} | \mathbf x_{1:\tau}, \mathbf w^{(n)}) }
\frac{ \partial p(\mathbf z_{1:m} | \mathbf x_{1:\tau}, \mathbf w^{(n)}) } { \partial y_k^t }.
\end{align}
Then, \eqref{eq:m_deriv} becomes
\begin{align}
\left. \frac{ \partial Q_\tau(\mathbf w | \mathbf x, \mathbf z, \mathbf w^{(n)}) } { \partial y_k^t } \right| _{\mathbf w = \mathbf w^{(n)}}  &=
\frac{1}{p(Z | \mathbf x_{1:\tau}, \mathbf w^{(n)})}
\sum_{m=0}^{|\mathbf z|}
\frac{ \partial p(\mathbf z_{1:m} | \mathbf x_{1:\tau}, \mathbf w^{(n)}) } { \partial y_k^t }
\\
&=
\frac{1}{p(Z | \mathbf x_{1:\tau}, \mathbf w^{(n)})}
\frac{\partial p(Z|\mathbf x_{1:\tau}, \mathbf w^{(n)})}{ \partial y_k^t}.
\label{eq:m_deriv2}
\end{align}
If we define the loss function to be minimized as
\begin{align}
\mathcal{L_\tau (\mathbf x, \mathbf z)} = -\ln p(Z | \mathbf x_{1:\tau}),
\end{align}
then its derivative equals to \eqref{eq:m_deriv2} with the opposite sign:
\begin{align}
\frac{ \partial \mathcal{L_\tau (\mathbf x, \mathbf z)}}
{\partial y_k^t}
=
-\left. \frac{ \partial Q_\tau(\mathbf w | \mathbf x, \mathbf z, \mathbf w^{(n)}) } { \partial y_k^t } \right| _{\mathbf w = \mathbf w^{(n)}} .
\end{align}
From now on, we drop $\mathbf w^{(n)}$ without loss of generality. Let
\begin{align}
\beta_{\tau, m}(\tau, u) =
\begin{cases}
1&\text{if } u=2m, 2m+1\\
0&\text{otherwise}
\end{cases} .
\end{align}
Following the standard CTC forward-backward equations in \citet{graves2012supervised},
\begin{align}
p(\mathbf z_{1:m} | \mathbf x_{1:\tau}) =
\sum_{u=1}^{|\mathbf z'|}
\alpha(t, u)\beta_{\tau, m}(t, u)
\label{eq:z_m}
\end{align}
\begin{align}
\frac{\partial p(\mathbf z_{1:m} | \mathbf x_{1:\tau})}
{\partial y_k^t} =
\frac{1}{y_k^t}
\sum_{u \in B( \mathbf z, k) }
\alpha(t, u)\beta_{\tau, m}(t, u).
\label{eq:z_m_delta}
\end{align}

From \eqref{eq:z_m} and \eqref{eq:z_m_delta}, $p(Z | \mathbf x_{1:\tau})$ and its derivative become
\begin{align}
p(Z | \mathbf x_{1:\tau}) =
\sum_{m=0}^{|\mathbf z|}
p(\mathbf z_{1:m} | \mathbf x_{1:\tau}) 
=
\sum_{u=1}^{|\mathbf z'|}
\alpha(t, u)\beta_{\tau}(t, u)
\end{align}
\begin{align}
\frac{\partial p(Z | \mathbf x_{1:\tau})}
{\partial y_k^t} =
\sum_{m=0}^{|\mathbf z|}
\frac{\partial p(\mathbf z_{1:m} | \mathbf x_{1:\tau})}
{\partial y_k^t} 
=
\frac{1}{y_k^t}
\sum_{u \in B( \mathbf z, k) }
\alpha(t, u)
\beta_{\tau}(t, u),
\end{align}
where the new backward variable for $p(Z | \mathbf x_{1:\tau})$ is
\begin{align}
\beta_{\tau}(t, u) =
\sum_{m=0}^{|\mathbf z|}
\beta_{\tau, m}(t, u),
\end{align}
which results in the simple initialization as
\begin{align}
\beta_{\tau}(\tau, u) = 1, \; \forall u.
\end{align}

Then, the error gradients become
\begin{align}
\frac{ \partial \mathcal{L_\tau (\mathbf x, \mathbf z)}}
{\partial y_k^t}
&=
-\frac{1}{p(Z | \mathbf x_{1:\tau})}
\frac{1}{y_k^t}
\sum_{u \in B( \mathbf z, k) }
\alpha(t, u)
\beta_{\tau}(t, u)\\
\frac{ \partial \mathcal{L_\tau (\mathbf x, \mathbf z)}}
{\partial a_k^t}
&=
y_k^t - \frac {1}{p(Z | \mathbf x_{1:\tau})} \sum_{u \in B(\mathbf z, k)} {\alpha(t, u) \beta_\tau(t, u)} ,
\end{align}
where
\begin{align}
p(Z | \mathbf x_{1:\tau})&=
\sum_{u=1}^{|\mathbf z'|}
\alpha(\tau, u) .
\end{align}

\subsection{Phoneme recognition on TIMIT}
\label{sec:timit}

\subsubsection{TIMIT corpus}

\begin{figure}[h]
\centering
\centerline{%
\begin{tikzpicture}
\begin{axis}
[
width=\columnwidth,
height=0.32\columnwidth,
bar width=0.04\columnwidth,
compat=1.3,
xmin=0,
ymin=0,
xmax=800,
ymax=1000,
label style={font=\footnotesize},
xlabel={Length of the sequence (frames)},
ylabel={Number of sequences},
xlabel shift=-2pt,
ylabel shift=-2pt,
ybar legend,
legend style={font=\scriptsize,at={(0.98,0.94)},anchor=north east},
tick label style={font=\scriptsize},
ymajorgrids,
minor x tick num=4,
minor y tick num=4,
xtick pos=both,
xtick align=inside,
major tick style={line width=0.010cm, black},
major tick length=0.10cm
]%
\legend{TIMIT training set (SA removed)};
\addplot[ybar, fill=cyan]
file{data/hist_timit.txt};
\end{axis}%
\end{tikzpicture}%
}%
\caption{Histogram of the length of the sequences in the TIMIT training set (SA removed), where the feature frames are extracted with the period of 10 ms.}
\label{fig:hist_timit}
\end{figure}
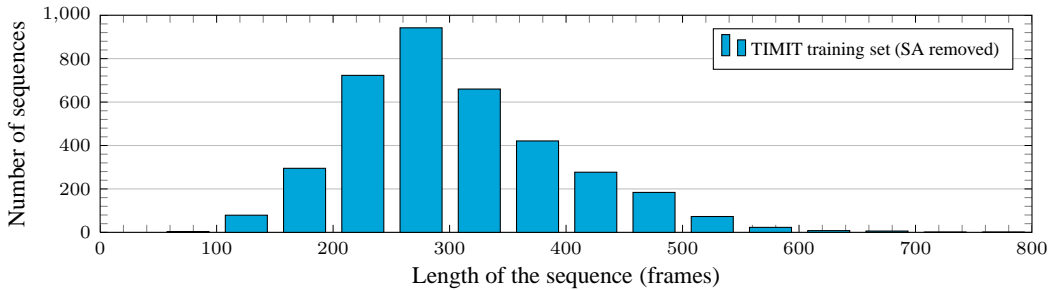

The TIMIT corpus \citep{garofolo1993darpa} contains American English recordings of 630 speakers from 8 major dialect regions in the United States. The training set contains about 3.1 hours of 3,696 utterances from 462 speakers after removing the SA recordings, in which only two sentences are spoken by multiple speakers. \figurename~\ref{fig:hist_timit} shows the histogram of the length of the training sequences, where the feature frames are extracted with the 10 ms period. The average length of the training sequences is 304 frames. We use the \emph{core test} set with 192 utterances as the test set. The development set contains the remaining 1,152 utterances that are obtained by excluding the \emph{core test} set from the \emph{complete test} set. The corpus also includes the full phonetic transcriptions.

\subsubsection{Network structure}

The network structure is a deep unidirectional LSTM RNN with 3 LSTM hidden layers, where each LSTM layer has 512 cells. The input is the same log Mel-frequency filterbank feature as in the WSJ experiments. The training procedure is also similar. The original TIMIT transcriptions are based on 61 phonetic labels. Accordingly, the RNN output is a 62-dimensional vector that consists of the probabilities of the original 61 phonemes and the extra CTC label. However, after decoding, they are mapped to 39 phonemes for evaluation as in \citet{lee1989speaker}.

\subsubsection{Training procedure}

For the experiments, the continuous CTC($2h'$;~$h'$) algorithm is employed so that the resulting RNN can run continuously on a infinitely long stream of the input speech. The networks are pre-trained with ADADELTA \citep{zeiler2012adadelta}, where the local learning rates are adaptively adjusted using the statistics of the recent gradient values. Before the online CTC training with the unroll amount greater than of equal to 512, the pre-training is performed for the 8 M ($8 \times 2^{20}$) training frames with the unroll amount of 2,048, the learning rate of $10^{-5}$, the Nesterov momentum of 0.9, and the RMS decay rate of $0.99$ for ADADELTA. On the other hand, we pre-trained the network with 12 M frames for the subsequent CTC training with less than 512 unroll steps. Unlike in the WSJ experiments, it is observed that applying the standard SGD method at the beginning often fails to initiate the training. We consider this is because the gradient computed by the SGD method is initially not noisy enough to help the parameters escape from the initial saddle point.

After the pre-training, the standard SGD is applied with the Nesterov momentum of 0.9. The training starts with the learning rate of $10^{-4}$. The intermediate evaluations are performed at every 2 M ($2 \times 2^{20}$) training frames on the development set with the best path decoding. If the phoneme error rate (PER) fails to improve during 6 consecutive evaluations, the learning rate decreases by the factor of 2 and the parameters are restored to those of the second best network. The training finishes when the learning rate becomes less than $10^{-6}$.

The network is regularized with dropout \citep{hinton2012improving} in both the pre-training and the main training stages following the approach in \citet{zaremba2014recurrent}, that is, dropout is only applied on the non-recurrent connections. The dropout rate is fixed to 0.5 throughout the experiments.

\subsubsection{Evaluation}

\begin{table}[t]
\caption{Comparison of CTC-TR coverages and PERs on the test set after CTC($2h'$;~$h'$) training with the varying amounts of unrolling}
\label{tbl:timit_per}
\begin{center}
\begin{tabular}{l|cc|cccc}
\multicolumn{1}{c|}{\bf \# Streams}  &\multicolumn{2}{c|}{\bf CTC-TR coverage (\%)} &\multicolumn{4}{c}{\bf PER (\%)} \\
\multicolumn{1}{c|}{\bf $\times$ \# Unroll}&\multicolumn{1}{c}{Average} &\multicolumn{1}{c|}{Maximum} &\multicolumn{1}{c}{Mean} &\multicolumn{1}{c}{$\pm$ Stdev.}& \multicolumn{1}{c}{Min. } &\multicolumn{1}{c}{Max.}
\\  \hline\hline \rule{0pt}{2.5ex}%
\hphantom{00}8 $\times$ 2,048 &100.0 &100.0 &21.14 &$\pm$ 0.29 &20.91 &21.57\\
\hphantom{0}16 $\times$ 1,024 &99.80 &100.0 &20.82 &$\pm$ 0.17 &20.66 &21.03 \\
\hphantom{0}32 $\times$ 512 &89.48 &99.60 &21.18 &$\pm$ 0.40 &20.60 &21.48\\
\hphantom{0}64 $\times$ 256 &60.69 &79.37 &20.77 &$\pm$ 0.24 &20.47 &20.97\\
128 $\times$ 128 &31.53 &42.02 &\bf{20.73} &$\pm$ 0.40 &\bf{20.39} &21.25\\
256 $\times$ 64 &15.77 &21.03 &21.00 &$\pm$ 0.16 &20.78 &21.15\\
\end{tabular}
\end{center}
\end{table}

\begin{table}[t]
\caption{Comparison of the proposed online CTC algorithm and the other models in the literature in terms of PER on the test set}
\label{tbl:timit_compare}
\begin{center}
\begin{minipage}{\textwidth}
\begin{center}
\begin{tabular}{l|c|c|c|c}
\multicolumn{1}{c|}{\bf Model}  &\multicolumn{1}{c|}{\bf Network (\# param)}  &\multicolumn{1}{c|}{\bf Bi-}  &\multicolumn{1}{c|}{\bf Test sequence} &\multicolumn{1}{c}{\bf PER (\%)}
\\  \hline\hline \rule{0pt}{2.5ex}
Proposed online CTC & LSTM (5.5 M) & {\bf No} & {\bf Almost infinite stream \footnote{Generated by concatenating all of the 192 test utterances}} & 20.73
\\  \hline \rule{0pt}{2.0ex}
\multirow{2}{*}{Attention-based model \footnote{\citet{chorowski2015attention}}} & \multirow{2}{*}{Conv.\footnote{Convolutional features}+GRU\footnote{Gated recurrent unit \citep{cho2014learning}}} & \multirow{2}{*}{\bf Yes} & {\bf Long sequences \footnote{Generated by concatenating 11 utterances}} & About 20 \\
& & & Utterance-wise & 17.6
\\  \hline \rule{0pt}{2.0ex}
RNN transducer \footnote{\label{graves}\citet{graves2013speech}}& LSTM (4.3 M) & Yes & Utterance-wise & 17.7
\\  \hline \rule{0pt}{2.0ex}
\multirow{2}{*}{Sequence-wise CTC \footref{graves}}  & \multirow{2}{*}{LSTM (3.8 M)} & Yes & \multirow{2}{*}{Utterance-wise} & 18.4 \\
& & No & & 19.6
\end{tabular}
\end{center}
\end{minipage}
\end{center}
\end{table}

The networks are evaluated on the very long test stream that is obtained by concatenating the entire test sequences. For the evaluation, the network output is decoded by the CTC beam search. The experiments are repeated 4 times and the mean and standard deviation estimates of PERs are reported based on the reduced 39-phoneme set.

The RNNs are unrolled 64, 128, 256, 512, 1,024, and 2,048 times. As shown in \tablename~\ref{tbl:timit_per}, the various unroll amounts make little difference to the final PERs on the test set. When the RNN is unrolled only 128 times, which is less than the average length of training sequences, the best PER of 20.73$\pm$0.40\% is obtained. On the other hand, the training with the unroll amount of 2,048 results in slightly degraded performance since it becomes harder for RNNs to catch the dependencies between the input and output sequences due to the noisy input frames from the consecutive sequences.

The performance of the proposed online CTC algorithm is compared with the other models in \tablename~\ref{tbl:timit_compare}. The other models employ early stopping to prevent overfitting and add weight noise while training for regularization. The bidirectional attention-based model in \citet{chorowski2015attention} shows 17.6\% PER with utterance-wise decoding. However, the PER increases to about 20\% with the long test sequences that are generated by concatenating 11 utterances. On the other hand, our CTC(128;~64)-trained unidirectional RNNs show 20.73$\pm$0.40\% PER with a very long test stream that is made by concatenating the entire 192 test utterances. Note that, unlike the CTC-trained unidirectional RNNs, the bidirectional models require unrolling in test time and have to listen the entire speech before generating outputs. Therefore, the proposed unidirectional RNN models are more suitable for realtime low-latency speech recognition systems without sacrificing much performance.

\end{document}